\documentclass[runningheads]{llncs}
\usepackage{amsfonts}
\usepackage{graphicx}
\usepackage{epsfig}
\usepackage{graphicx}
\usepackage{amsmath}
\usepackage{amssymb}
\usepackage{mathtools}
\usepackage{resizegather}
\usepackage{marvosym}
\usepackage[linesnumbered,ruled]{algorithm2e}
\usepackage{dsfont}
\usepackage{enumitem}
\usepackage{caption}
\usepackage{multirow}
\usepackage{setspace}

\begin{document}
\title{Generative Self-training for Cross-domain Unsupervised Tagged-to-Cine MRI Synthesis}
\titlerunning{Generative Self-training for Tagged-to-Cine UDA}
%
 
%
\author{Xiaofeng Liu\inst{1} \and Fangxu Xing\inst{1} \and Maureen Stone\inst{2} \and Jiachen Zhuo\inst{2} \and Timothy Reese\inst{3} \and Jerry L. Prince\inst{4} \and Georges El Fakhri\inst{1} \and Jonghye Woo\inst{1}}


\institute{Gordon Center for Medical Imaging, Department of Radiology, Massachusetts General Hospital and Harvard Medical School, Boston, MA, 02114, USA\and
Dept. of Neural and Pain Sciences, University of Maryland School of Dentistry, Baltimore, MD, USA\and
Athinoula A. Martinos Center for Biomedical Imaging, Dept. of Radiology, Massachusetts General Hospital and Harvard Medical School, Boston, MA, USA\and
Dept. of Electrical and Computer Engineering, Johns Hopkins University, Baltimore, MD, USA}

\authorrunning{X. Liu et al.}
\maketitle              

\begin{abstract}
Self-training based unsupervised domain adaptation (UDA) has shown great potential to address the problem of domain shift, when applying a trained deep learning model in a source domain to unlabeled target domains. However, while the self-training UDA has demonstrated its effectiveness on discriminative tasks, such as classification and segmentation, via the reliable pseudo-label selection based on the softmax discrete histogram, the self-training UDA for generative tasks, such as image synthesis, is not fully investigated. In this work, we propose a novel generative self-training (GST) UDA framework with continuous value prediction and regression objective for cross-domain image synthesis. Specifically, we propose to filter the pseudo-label with an uncertainty mask, and quantify the predictive confidence of generated images with practical variational Bayes learning. The fast test-time adaptation is achieved by a round-based alternative optimization scheme. We validated our framework on the tagged-to-cine magnetic resonance imaging (MRI) synthesis problem, where datasets in the source and target domains were acquired from different scanners or centers. Extensive validations were carried out to verify our framework against popular adversarial training UDA methods. Results show that our GST, with tagged MRI of test subjects in new target domains, improved the synthesis quality by a large margin, compared with the adversarial training UDA methods.

\end{abstract}

\section{Introduction}

Deep learning has advanced state-of-the-art machine learning approaches and excelled at learning  representations suitable for numerous discriminative and generative tasks \cite{wang2021automated,liu2018ordinal,liu2020unimodal,liu2021symmetric}. However, a deep learning model trained on labeled data from a source domain, in general, performs poorly on unlabeled data from unseen target domains, partly because of discrepancies between source and target data distributions, i.e., domain shift \cite{liu2021Generalization}. The problem of domain shift in medical imaging arises, because data are often acquired from different scanners, protocols, or centers \cite{liu2021subtype}. This issue has motivated many researchers to investigate unsupervised domain adaptation (UDA), which aims to transfer knowledge learned from a labeled source domain to different but related unlabeled target domains \cite{wang2018deep,zou2019confidence}.  

There has been a great deal of work to alleviate the domain shift using UDA \cite{wang2018deep}. Early methods attempted to learn domain-invariant representations or to take instance importance into consideration to bridge the gap between the source and target domains. In addition, due to the ability of deep learning to disentangle explanatory factors of variations, efforts have been made to learn more transferable features. Recent works in UDA incorporated discrepancy measures into network architectures to align feature distributions between source and target domains \cite{liu2021Off-the-Shelf,liu2021unified}. This was achieved by either minimizing the distribution discrepancy between feature distribution statistics, e.g., maximum mean discrepancy (MMD), or adversarially learning the feature representations to fool a domain classifier in a two-player minimax game~\cite{liu2021Off-the-Shelf}.

Recently, self-training based UDA presents a powerful means to counter unknown labels in the target domain \cite{zou2019confidence}, surpassing the adversarial learning-based methods in many discriminative UDA benchmarks, e.g., classification and segmentation (i.e., pixel-wise classification)~\cite{wei2021theoretical,mei2020instance,shin2020two}. The core idea behind the deep self-training based UDA is to iteratively generate a set of one-hot (or smoothed) pseudo-labels in the target domain, followed by retraining the network based on these pseudo-labels with target data \cite{zou2019confidence}. Since outputs of the previous round can be noisy, it is critical to only select the high confidence prediction as reliable pseudo-label. In discriminative self-training with softmax output unit and cross-entropy objective, it is natural to define the confidence for a sample as the max of its output softmax probabilities \cite{zou2019confidence}. Calibrating the uncertainty of the regression task, however, can be more challenging. Because of the insufficient target data and unreliable pseudo-labels, there can be both \textit{epistemic} and \textit{aleatoric} uncertainties \cite{der2009aleatory} in self-training UDA. In addition, while the self-training UDA has demonstrated its effectiveness on classification and segmentation, via the reliable pseudo-label selection based on the softmax discrete histogram, the same approach for generative tasks, such as image synthesis, is underexplored.

In this work,  we propose a novel generative self-training (GST) UDA framework with continuous value prediction and regression objective for tagged-to-cine magnetic resonance (MR) image synthesis. More specifically, we propose to filter the pseudo-label with an uncertainty mask, and quantify the predictive confidence of generated images with practical variational Bayes learning. The fast test-time adaptation is achieved by a round-based alternative optimization scheme. Our contributions are summarized as follows:

\noindent$\bullet$ We propose to achieve cross-scanner and cross-center test-time UDA of tagged-to-cine MR image synthesis, which can potentially reduce the extra cine MRI acquisition time and cost.

\noindent$\bullet$ A novel GST UDA scheme is proposed, which controls the confident pseudo-label (continuous value) selection with a practical Bayesian uncertainty mask. Both the aleatoric and epistemic uncertainties in GST UDA are investigated.   

\noindent$\bullet$ Both quantitative and qualitative evaluation results, using a total of 1,768 paired slices of tagged and cine MRI from the source domain and tagged MR slices of target subjects from the cross-scanner and cross-center target domain, demonstrate the validity of our proposed GST framework and its superiority to conventional adversarial training based UDA methods.

\begin{figure}[t]
\begin{center}
\includegraphics[width=1\linewidth]{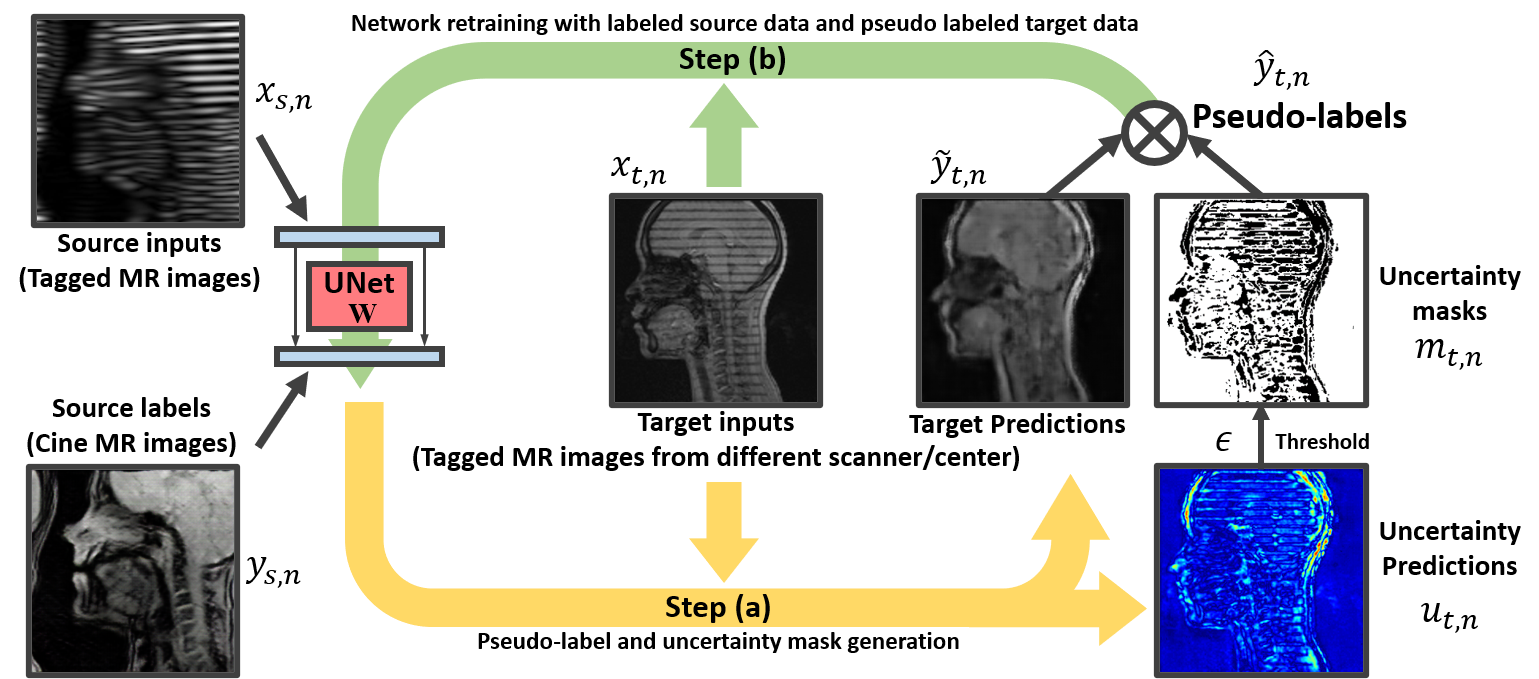}
\end{center} 
\caption{Illustration of our generative self-training UDA for tagged-to-cine MR image synthesis. In each iteration, two-step alternative training is carried out.} 
\label{ccc}\end{figure}

\section{Methodology} 

In our setting of the UDA image synthesis, we have paired resized tagged MR images, $\mathbf{x}_s\in\mathbb{R}^{256\times256}$, and cine MR images, $\mathbf{y}_s\in\mathbb{R}^{256\times256}$, indexed by $s=1, 2,\cdots,S $, from the source domain $\{\mathbf{X}_S, \mathbf{Y}_S\}$, and target samples $\mathbf{x}_t\in\mathbb{R}^{256\times256}$ from the unlabeled target domain $\mathbf{X}_T$, indexed by $t=1,2,\cdots, T$. In both training and testing, the ground-truth target labels, i.e., cine MR images in the target domain, are inaccessible, and the pseudo-label $\hat{\mathbf{y}}_t\in\mathbb{R}^{256\times256}$ of $\mathbf{x}_t$ is iteratively generated in a self-training scheme \cite{zou2019confidence,liu2020energy}. In this work, we adopt the U-Net-based Pix2Pix \cite{isola2017image} as our translator backbone, and initialize the network parameters $\mathbf{w}$ with the pre-training using the labeled source domain $\{\mathbf{X}_S, \mathbf{Y}_S\}$. In what follows, alternative optimization based self-training is applied to gradually update the U-Net part for the target domain image synthesis by training on both $\{\mathbf{X}_S, \mathbf{Y}_S\}$ and $\mathbf{X}_T$. Fig. \ref{ccc} illustrates the proposed algorithm flow, which is detailed below.

\subsection{Generative Self-training UDA} 

The conventional self-training regards the pseudo-label $\hat{\mathbf{y}}_{t}$ as a learnable latent variable in the form of a categorical histogram, and assigns all-zero vector label for the uncertain samples or pixels to filter them out for loss calculation \cite{zou2019confidence,liu2020energy}. Since not all pseudo-labels are reliable, we define a confidence threshold to progressively select confident pseudo-labels \cite{zhu2007semi}. This is akin to self-paced learning that learns samples in an easy-to-hard order \cite{kumar2010self,tang2012shifting}. In classification or segmentation tasks, the confidence can be simply measured by the maximum softmax output histogram probability \cite{zou2019confidence}. The output of a generation task, however, is continuous values and thus setting the pseudo-label as 0 cannot drop the uncertain sample in the regression loss calculation.

Therefore, we first propose to formulate the generative self-training as a unified regression loss minimization scheme, where pseudo-labels can be a pixel-wise continuous value and indicate the uncertain pixel with an uncertainty mask $\mathbf{m}_{t}=\{{m}_{t,n}\}_{n=1}^{256\times256}$, where $n$ indexes the pixel in the images, and ${{m}}_{t,n}\in \{0,1\},\forall t,n$: 
\begin{align}\label{111}
&\begin{matrix}\underset{\mathbf{w},\mathbf{m}_t}{\mathop{\min}}~~\underbrace{\sum\limits_{{{s}}\in {{S}}}{\sum\limits_{n=1}^{N}}  ||y_{s,n}-\tilde{y}_{s,n}||^2_2} + \\ {\mathcal{L}_{reg}^{s}(\mathbf{w})} \end{matrix}\begin{matrix}\underbrace{\sum\limits_{{{t}}\in {{T}}}{\sum\limits_{n=1}^{N}}  ||(\hat{y}_{t,n}-\tilde{y}_{t,n})m_{t,n}||^2_2}\\ {\mathcal{L}_{reg}^{t}(\mathbf{w},\mathbf{m}_t)}\end{matrix}  \\
&~~s.t. ~~m_{t,n}=\begin{cases}1 & u_{t,n} < \epsilon\\0 & u_{t,n} > \epsilon \end{cases}; ~~\epsilon=\text{min}\{\text{top}~p\%~\text{sorted}~u_{t,n}\},
\end{align} where ${{x}}_{s,n},{{y}}_{s,n},{{x}}_{t,n},\hat{{y}}_{t,n}\in [0,255]$. For example, $y_{s,n}$ indicates the $n$-th pixel of the $s$-th source domain ground-truth cine MR image ${\mathbf{y}}_{s}$. $\tilde{y}_{s,n}$ and $\tilde{y}_{t,n}$ represent the generated source and target images, respectively. $\mathcal{L}_{reg}^{s}(\mathbf{w})$ and $\mathcal{L}_{reg}^{t}(\mathbf{w},\mathbf{m}_t)$ are the regression loss of the source and target domain samples, respectively. Notably, there is only one network parameterized with $\mathbf{w}$, which is updated with the loss in both domains. ${{u}}_{t,n}$ is the to-be estimated uncertainty of a pixel and determines the value of the uncertainty mask ${{m}}_{t,n}$ with a threshold $\epsilon$. $\epsilon$ is a critical parameter to control pseudo-label learning and selection, which is determined by a single meta portion parameter $p$, indicating the portion of pixels to be selected in the target domain. Empirically, we define $\epsilon$ in each iteration, by sorting ${{u}}_{t,n}$ in increasing order and set $\epsilon$ to minimum ${{u}}_{t,n}$ of the top $p\in[0,1]$ percentile rank.



\subsection{Bayesian Uncertainty Mask for Target Samples} 

Determining the mask value ${{m}}_{t,n}$ for the target sample requires the uncertainty estimation of ${{u}}_{t,n}$ in our self-training UDA. Notably, the lack of sufficient target domain data can result in the $epistemic$ uncertainty w.r.t. the model parameters, while the noisy pseudo-label can lead to the $aleatoric$ uncertainty \cite{der2009aleatory,kendall2017uncertainties,hu2019supervised}. 

To counter this, we model the $epistemic$ uncertainty via Bayesian neural networks which learn a posterior distribution $p(\mathbf{w}|\mathbf{X}_T,\mathbf{\hat{Y}}_T)$ over the probabilistic model parameters rather than a set of deterministic parameters \cite{rasmussen2003gaussian}. In particular, a tractable solution is to replace the true posterior distribution with a variational approximation $q(\mathbf{w})$, and dropout variational inference can be a practical technique. This can be seen as using the Bernoulli distribution as the approximation distribution $q(\mathbf{w})$ \cite{gal2015bayesian}. The $K$ times prediction with independent dropout sampling is referred to as Monte Carlo (MC) dropout. We use the mean squared error (MSE) to measure the epistemic uncertainty as in \cite{rasmussen2003gaussian}, which assesses a one-dimensional regression model similar to \cite{fruehwirt2018bayesian}. Therefore, the epistemic uncertainty with MSE of each pixel with $K$ times dropout generation is given by
\begin{align}\label{222} 
u^{epistemic}_{t,n} = \frac{1}{K} \sum_{k=1}^{K} || {\tilde{y}_{t,n}-\mu_{t,n}}||^2_2 ; ~~\mu_{t,n}=\frac{1}{K} \sum_{k=1}^{K} \tilde{y}_{t,n}, 
\end{align} 
where $\mu_{t,n}$ is the predictive mean of $\tilde{y}_{t,n}$. 


Because of the different hardness and divergence and because the pseudo-label noise can vary for different $\mathbf{x}_t$, the heteroscedastic $aleatoric$ uncertainty modeling is required \cite{nix1994estimating,le2005heteroscedastic}. In this work, we use our network to transform $\mathbf{x}_t$, with its head split to predict both $\tilde{\mathbf{y}}_t$ and the variance map $\mathbf{\sigma}^2_t\in\mathbb{R}^{256\times256}$; and its element $\sigma^2_{t,n}$ is the predicted variance for the $n$-th pixel. We do not need ``uncertainty labels" to learn $\mathbf{\sigma}^2_t$ prediction. Rather, we can learn $\mathbf{\sigma}^2_t$ implicitly from a regression loss function \cite{le2005heteroscedastic,kendall2017uncertainties}. The masked regression loss can be formulated as 
\begin{align}\label{333}\mathcal{L}_{reg}^{t}(\mathbf{w},\mathbf{m}_t,\sigma^2_{t})= \sum\limits_{{{t}}\in {{T}}}{\sum\limits_{n=1}^{N}}  (\frac{1}{\sigma^2_{t,n}}||(\hat{y}_{t,n}-\tilde{y}_{t,n})m_{t,n}||^2_2+ \beta\text{log} \sigma^2_{t,n}),\end{align} which consists of a variance normalized residual regression term and an uncertainty regularization term. The second regularization term keeps the network from predicting an infinite uncertainty, i.e., zero loss, for all the data points. Then, the averaged aleatoric uncertainty of $K$ times MC dropout can be measured by $u^{aleatoric}_{t,n}=\frac{1}{K} \sum_{k=1}^{K}\sigma^2_{t,n}$ \cite{le2005heteroscedastic,kendall2017uncertainties}.

Moreover, minimizing Eq.~(\ref{333}) can be regarded as the Lagrangian with a multiplier $\beta$ of $\underset{\mathbf{w}}{\mathop{\min }}\sum\limits_{{{t}}\in {{T}}}{\sum\limits_{n=1}^{N}}  \frac{1}{\sigma^2_{t,n}}||(\hat{y}_{t,n}-\tilde{y}_{t,n})m_{t,n}||^2_2;~s.t.~\sum\limits_{{{t}}\in {{T}}}{\sum\limits_{n=1}^{N}} \text{log} \sigma^2_{t,n}< C$\footnote{It can be rewritten as $\underset{\mathbf{w}}{\mathop{\min}}~ \mathcal{F}=\{\sum\limits_{{{t}}\in {{T}}}{\sum\limits_{n=1}^{N}}  \frac{1}{\sigma^2_{t,n}}||(\hat{y}_{t,n}-\tilde{y}_{t,n})m_{t,n}||^2_2+\beta(\sum\limits_{{{t}}\in {{T}}}{\sum\limits_{n=1}^{N}} \text{log} \sigma^2_{t,n}-C)\}$. Since $\beta,C\geq0$, an upper bound on $\mathcal{F}$ can be obtained as $\mathcal{F}\leq \mathcal{L}_{reg}^t$.}, where $C\in\mathbb{R}^+$ indicates the strength of the applied constraint. The condition term essentially controls the target domain predictive uncertainty, which is helpful for UDA \cite{han2019unsupervised}. Our final pixel-wise self-training UDA uncertainty $u_{t,n}=u^{epistemic}_{t,n}+u^{aleatoric}_{t,n}$ is a combination of the two uncertainties \cite{kendall2017uncertainties}.




\subsection{Training Protocol} 

As pointed out in \cite{grandvalet2006entropy}, directly optimizing the self-training objectives can be difficult and thus the deterministic annealing expectation maximization (EM) algorithms are often used instead. Specifically, the generative self-training can be solved by alternating optimization based on the following \textbf{a)} and \textbf{b)} steps.

\noindent \textbf{a) Pseudo-label and uncertainty mask generation.} \label{a)}~With the current $\mathbf{w}$, apply the MC dropout for $K$ times image translation of each target domain tagged MR image $\mathbf{x}_{t}$. We estimate the pixel-wise uncertainty $u_{t,n}$, and calculate the uncertainty mask ${\mathbf{m}}_t$ with the threshold $\epsilon$. We set the pseudo-label of the selected pixel in this round as $\hat{{y}}_{t,n}={{\mu}}_{t,n}$, i.e., the average value of $K$ outputs. 

\noindent \textbf{b) Network $\mathbf{w}$ retraining}. \label{b)}~Fix $\hat{\mathbf{Y}}_T=\{\hat{\mathbf{y}}_t\}_{t=1}^{T}$, ${\mathbf{M}}_T=\{{\mathbf{m}}_t\}_{t=1}^{T}$ and solve:
\begin{align}\label{st_b}
& \underset{\mathbf{w}}{\mathop{\min }}~~ \sum\limits_{{{s}}\in {{S}}}{\sum\limits_{n=1}^{N}}  ||y_{s,n}-\tilde{y}_{s,n}||^2_2  +   \sum\limits_{{{t}}\in {{T}}}{\sum\limits_{n=1}^{N}}  (\frac{1}{\sigma^2_{t,n}}||(\hat{y}_{t,n}-\tilde{y}_{t,n})m_{t,n}||^2_2+ \beta\text{log} \sigma^2_{t,n}) 
\end{align} to update $\mathbf{w}$. Carrying out step \textbf{a)} and \textbf{b)} for one time is defined as one round in self-training. Intuitively, step \textbf{a)} is equivalent to simultaneously conducting pseudo-label learning and selection. In order to solve step \textbf{b)}, we can use a typical gradient method, e.g. Stochastic Gradient Descent (SGD). The meta parameter $p$ is linearly increasing from 30\% to 80\% alongside the training to incorporate more pseudo-labels in the subsequent rounds as in \cite{zou2019confidence}.

\section{Experiments and Results} 

\begin{figure*}[t]
\begin{center} 
\includegraphics[width=0.98\linewidth]{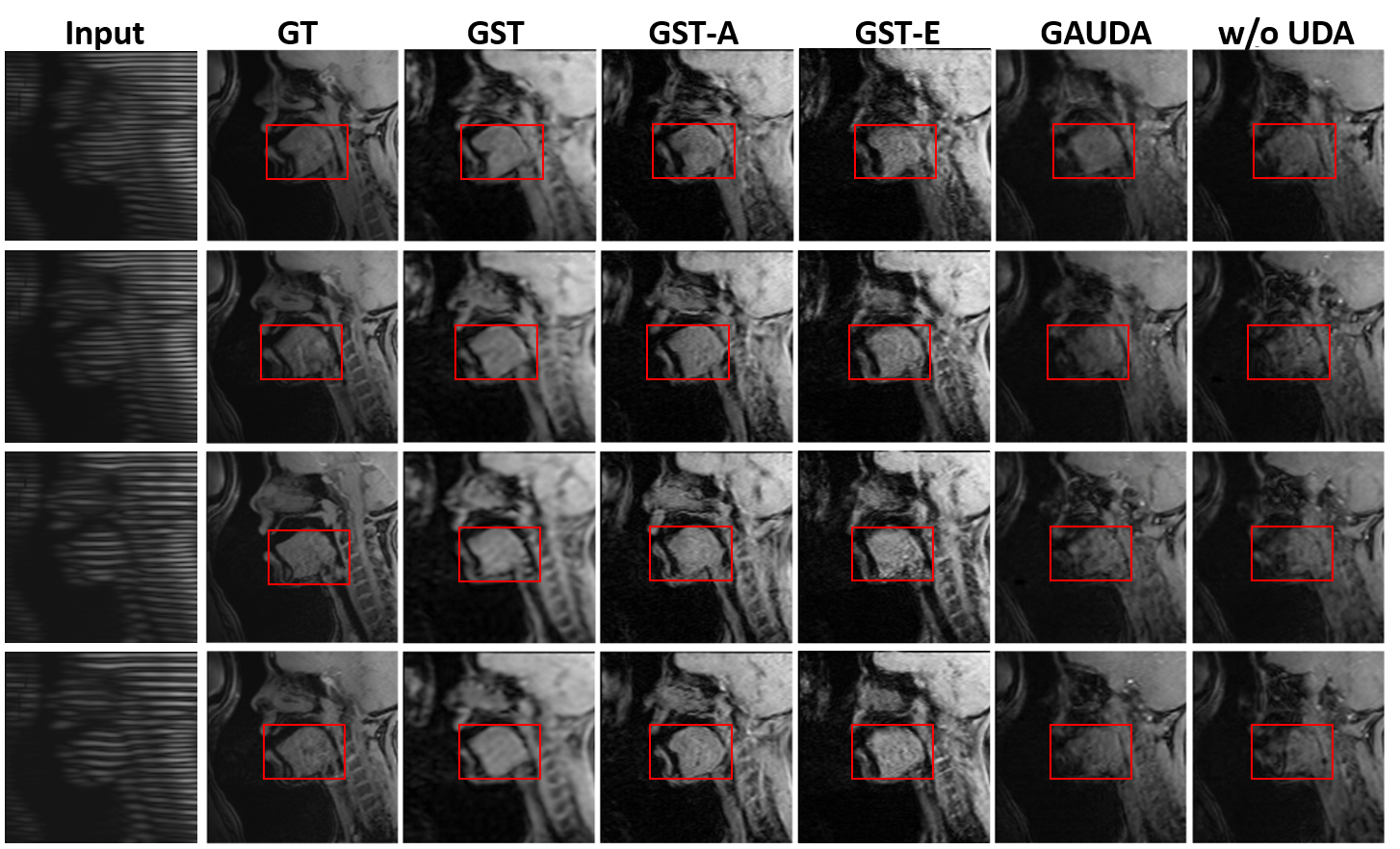} 
\end{center} 
\caption{Comparison of different UDA methods on the cross-scanner tagged-to-cine MR image synthesis task, including our proposed GST, GST-A, and GST-E, adversarial UDA \cite{cui2020gradually}*, and Pix2Pix~\cite{isola2017image} without adaptation. * indicates the first attempt at tagged-to-cine MR image synthesis. GT indicates the ground-truth.}
\label{fig:results1}
\end{figure*}

We evaluated our framework on both cross-scanner and cross-center tagged-to-cine MR image synthesis tasks. For the labeled source domain, a total of 1,768 paired tagged and cine MR images from 10 healthy subjects at clinical center A were acquired. We followed the test time UDA setting \cite{karani2021test}, which uses only one unlabeled target subject in UDA training and testing. 

For fair comparison, we adopted Pix2Pix \cite{isola2017image} for our source domain training as in \cite{liu2021dual}, and used the trained U-Net as the source model for all of the comparison methods. In order to align the absolute value of each loss, we empirically set weight $\beta=1$ and $K=20$. Our framework was implemented using the PyTorch deep learning toolbox. The GST training was performed on a V100 GPU, which took about 30 min. We note that $K$ times MC dropout can be processed parallel. In each iteration, we sampled the same number of source and target domain samples. 

\begin{table}[t]
\centering
\caption{Numerical comparisons of cross-scanner and cross-center evaluations. $\pm$ standard deviation is reported over three evaluations.} 
\resizebox{1\linewidth}{!}{
\begin{tabular}{c|c|c|c|c||ccc|c|cc}
\hline
 &\multicolumn{4}{c||}{Cross-scanner} & Cross-center\\ \hline

Methods& L1~$\downarrow$& SSIM~$\uparrow$ &	PSNR~$\uparrow$ &  IS~$\uparrow$&  IS~$\uparrow$\\\hline\hline
w/o UDA  \cite{isola2017image}     	& {176.4}$\pm$0.1&	0.8325$\pm$0.0012&	26.31$\pm$0.05&	8.73$\pm$0.12  &	5.32$\pm$0.11\\\hline
ADDA \cite{tzeng2017adversarial}	&168.2$\pm$0.2&	0.8784$\pm$0.0013&	33.15$\pm$0.04&	10.38$\pm$0.11& 8.69$\pm$0.10\\
GAUDA \cite{cui2020gradually}      	&161.7$\pm$0.1& {0.8813}$\pm$0.0012&	 {33.27}$\pm$0.06& {10.62}$\pm$0.13&	8.83$\pm$0.14\\\hline
GST                              &\textbf{158.6}$\pm$0.2&	\textbf{0.9078}$\pm$0.0011& \textbf{34.48}$\pm$0.05&  \textbf{12.63}$\pm$0.12&	\textbf{9.76}$\pm$0.11\\
GST-A                              & {159.5}$\pm$0.3&	 {0.8997}$\pm$0.0011&  {34.03}$\pm$0.04&   {12.03}$\pm$0.12&	9.54$\pm$0.13\\
GST-E                              & {159.8}$\pm$0.1&	 {0.9026}$\pm$0.0013&  {34.05}$\pm$0.05&   {11.95}$\pm$0.11&	9.58$\pm$0.12\\\hline
\end{tabular}
}
\label{tabel:1} 
\end{table}

\subsection{Cross-scanner tagged-to-cine MR image synthesis} 

In the cross-scanner image synthesis setting, a total of 1,014 paired tagged and cine MR images from 5 healthy subjects in the target domain were acquired at clinical center A with a different scanner. As a result, there was an appearance discrepancy between the source and target domains.

The synthesis results using source domain Pix2Pix \cite{isola2017image} without UDA training, gradually adversarial UDA (GAUDA) \cite{cui2020gradually}, and our proposed framework are shown in Fig.~\ref{fig:results1}. Note that GAUDA with source domain initialization took about 2 hours for the training, which was four times slower than our GST framework. In addition, it was challenging to stabilize the adversarial training \cite{che2021deep}, thus yielding checkerboard artifacts. Furthermore, the hallucinated content with the domain-wise distribution alignment loss produced a relatively significant difference in shape and texture within the tongue between the real cine MR images. By contrast, our framework achieved the adaptation with relatively limited target data in the test time UDA setting \cite{karani2021test}, with faster convergence time. In addition, our framework did not rely on adversarial training, generating visually pleasing results with better structural consistency as shown in Fig.~\ref{fig:results1}, which is crucial for subsequent analyses such as segmentation.

For an ablation study, in Fig.~\ref{fig:results1}, we show the performance of GST without the aleatoric or epistemic uncertainty for the uncertainty mask, i.e., GST-A or GST-E. Without measuring the aleatoric uncertainty caused by the inaccurate label, GST-A exhibited a small distortion of the shape and boundary. Without measuring the epistemic uncertainty, GST-E yielded noisier results than GST.

The synthesized images were expected to have realistic-looking textures, and to be structurally cohesive with their corresponding ground truth images. For quantitative evaluation, we adopted widely used evaluation metrics: mean L1 error, structural similarity index measure (SSIM), peak signal-to-noise ratio (PSNR), and unsupervised inception score (IS) \cite{liu2021dual}. Table \ref{tabel:1} lists numerical comparisons using 5 testing subjects. The proposed GST outperformed GAUDA \cite{cui2020gradually} and ADDA \cite{tzeng2017adversarial} w.r.t. L1 error, SSIM, PSNR, and IS by a large margin.
 
\subsection{Cross-center tagged-to-cine MR image synthesis} 
To further demonstrate the generality of our framework for the cross-center tagged-to-cine MR image synthesis task, we collected 120 tagged MR slices of a subject at clinical center B with a different scanner. As a result, the data at clinical center B had different soft tissue contrast and tag spacing, compared with clinical center A, and the head position was also different. 

\begin{figure*}[t]
\begin{center}
\includegraphics[width=0.97\linewidth]{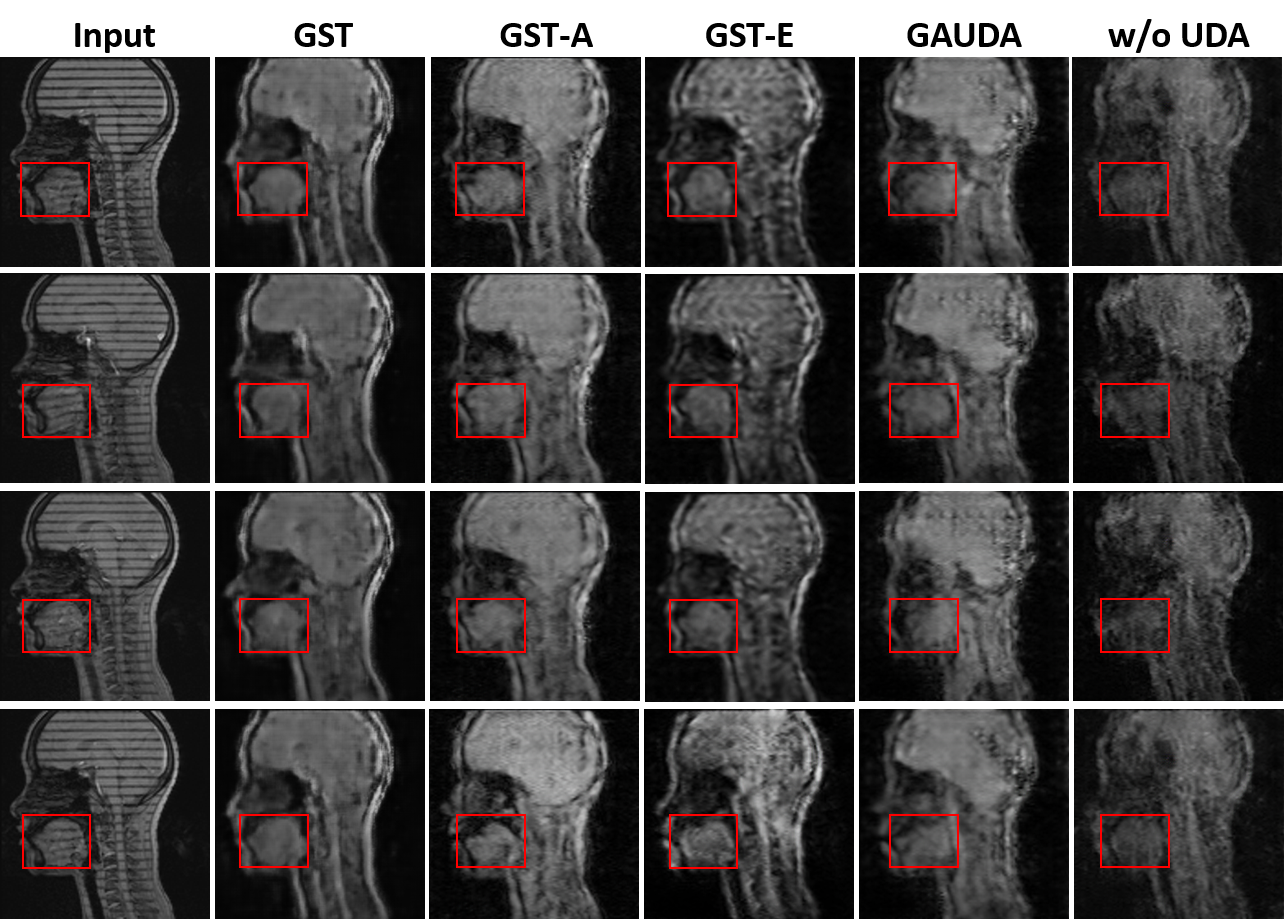} 
\end{center}  
\caption{Comparison of different UDA methods on the cross-center tagged-to-cine MR image synthesis task, including our proposed GST, GST-A, and GST-E, adversarial UDA \cite{cui2020gradually}*, and Pix2Pix~\cite{isola2017image} without adaptation. * indicates the first attempt at tagged-to-cine MR image synthesis.} 
\label{fig:results2}
\end{figure*}

The qualitative results in Fig.~\ref{fig:results2} show that the anatomical structure of the tongue is better maintained using our framework with both the aleatoric and epistemic uncertainties. Due to the large domain gap present in the datasets between the two centers, the overall synthesis quality was not as good as the cross-scanner image synthesis task, as visually assessed. In Table \ref{tabel:1}, we provide the quantitative comparison using IS, which does not need the paired ground truth cine MR images \cite{liu2021dual}. Consistently with the cross-scanner setting, our GST outperformed adversarial training methods, including GAUDA and ADDA \cite{cui2020gradually,tzeng2017adversarial}, indicating the self-training can be a powerful technique for the generative UDA task, similar to the conventional discriminative self-training \cite{zou2019confidence,liu2020energy}.

\section{Discussion and Conclusion} 

In this work, we presented a novel generative self-training framework for UDA and applied the framework to cross-scanner and cross-center tagged-to-MR image synthesis tasks. With a practical yet principled Bayesian uncertainty mask, our framework was able to control the confident pseudo-label selection. In addition, we systematically investigated both the aleatoric and epistemic uncertainties in generative self-training UDA. Our experimental results demonstrated that our framework yielded the superior performance, compared with the popular adversarial training UDA methods, as quantitatively and qualitatively assessed. The synthesized cine MRI with test time UDA can potentially be used to segment the tongue and to observe surface motion, without the additional acquisition cost and time.

\section*{Acknowledgements}

This work is supported by NIH R01DC014717, R01DC018511, and R01CA133015.

\bibliographystyle{splncs04}
\bibliography{egbib}

\end{document}